\documentclass[conference]{IEEEtran}
\IEEEoverridecommandlockouts
\usepackage{cite}
\usepackage{amsmath,amssymb,amsfonts}
\usepackage{algorithmic}
\usepackage{graphicx}
\usepackage{textcomp}
\usepackage{xcolor}
\def\BibTeX{{\rm B\kern-.05em{\sc i\kern-.025em b}\kern-.08em
    T\kern-.1667em\lower.7ex\hbox{E}\kern-.125emX}}
\begin{document}

\title{Logarithmic Loss Transformation to Improve Anomaly Detection Model Training Stability
}

\author{\IEEEauthorblockN{YeongHyeon Park}
\IEEEauthorblockA{\textit{IoT Solution Business Group} \\
\textit{SK Planet Co., Ltd.}\\
Seongnam, Republic of Korea \\
yeonghyeon@sk.com}
}

\maketitle

\begin{abstract}
Recently, deep learning-based algorithms are widely adopted due to the advantage of being able to establish anomaly detection models without or with minimal domain knowledge of the task. Instead, to train the artificial neural network more stable, it should be better to define the appropriate neural network structure or the loss function. For the training anomaly detection model, the mean squared error (MSE) function is adopted widely. On the other hand, the novel loss function, logarithmic mean squared error (LMSE), is proposed in this paper to train the neural network more stable. This study covers a variety of comparisons from mathematical comparisons, visualization in the differential domain for backpropagation, loss convergence in the training process, and anomaly detection performance. In an overall view, LMSE is superior to the existing MSE function in terms of strongness of loss convergence, anomaly detection performance. The LMSE function is expected to be applicable for training not only the anomaly detection model but also the general generative neural network.
\end{abstract}


\section{Introduction}
\label{sec:introduction}

Recently, many anomaly detection systems are constructed with generative neural networks such as an auto-encoder or generative adversarial network \cite{mse_aytekin2018norm, mse_oh2018resi, mse_park2018fared, mse_nguyen2019conad, mse_gong2019memae, mse_schlegl2019fanogan, mse_chow2020cae, mse_park2020hpgan}. For enabling the conventional rule-based anomaly detection system, establishing the detailed criteria for determining whether the current state is normal or not should be preceded \cite{rule_wong2002rule, rule_heinrich2020rule}. 

In the case of conventional machine learning algorithms, defining the delicate feature of the data is needed to train the model \cite{ml_chan2003feat, ml_ahmed2007feat, ml_khalid2014feat}. If the machine learning system operator has advanced knowledge or experience in the target problem, the feature information for the machine learning model training will be defined more sophisticatedly. However, there are limitations that it takes a lot of time, effort, and cost to become such a high-level operator. Also, the system dependence on the machine learning operator will be increased in this case.

\begin{figure}[t]
    \begin{center}
        \includegraphics[width=0.90\linewidth]{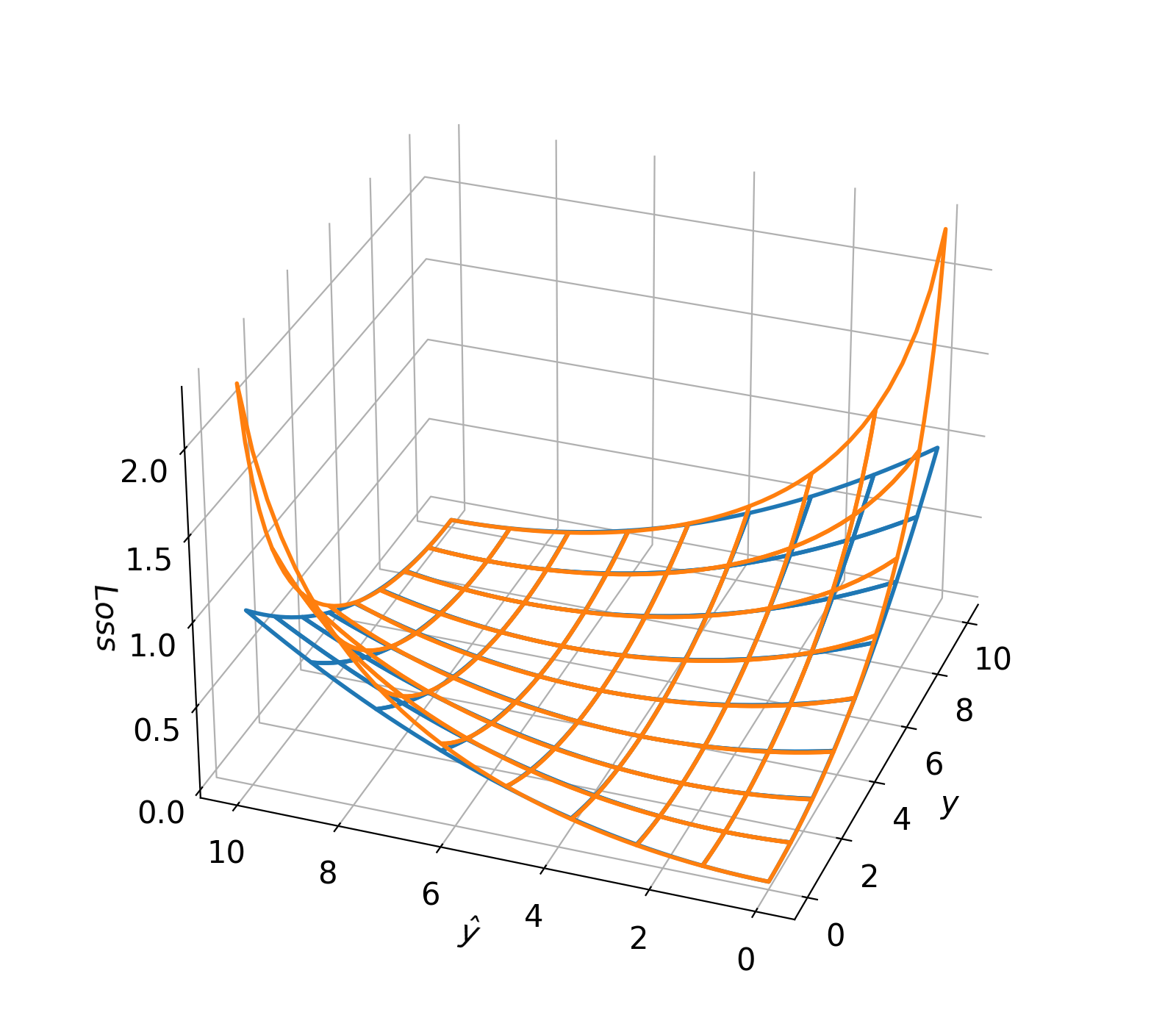}
    \end{center}
    \vspace*{-5mm}
    \caption{Visualization of the MSE and LMSE surfaces with blue and orange colors respectively. The LMSE surface shows a steeper gradient than the original MSE. Thus, we can eliminate the large error situation through the LMSE better than the MSE and ensure training stability.}
    \label{fig:surface}
\end{figure}

When using deep learning techniques, high-performance computing environments such as a graphic processing unit are required for training purposes. Vicariously, deep learning method can take over for automating the delicate tasks based on domain knowledge such as decision rule or feature definition in the aforementioned conventional method.

In order to train the neural network better, operators only need to define the model structure and loss function will a background in mathematics or computer science. Various effective loss functions and neural networks structures have already been proposed. The mean squared error (MSE) is generally used as both a target function in the training process and an indicator for determining normal or not. 

On the other hand, a novel concise loss function is proposed in this paper to enhance the robustness of the training procedure and anomaly detection performance. This study clarifies the effect of the proposed LMSE function from the process of constructing the loss function to the experimental comparison with the conventional MSE function.

\section{Related work}
This section describes related studies with anomaly detection and loss function before proposing the novel loss function.

\label{sec:introduction}

\subsection{Deep learning-based anomaly detection}
\label{subsec:dl_anomaly_detection}
Lots of generative neural network-based anomaly detection algorithms adopt the MSE as an indicator to decide normal or not \cite{mse_aytekin2018norm, mse_oh2018resi, mse_park2018fared, mse_nguyen2019conad, mse_gong2019memae, mse_schlegl2019fanogan, mse_chow2020cae, mse_park2020hpgan}. Among the above, some architectures use the MSE, as a target function for training \cite{mse_aytekin2018norm, mse_oh2018resi, mse_park2018fared, mse_chow2020cae}. When using MSE for both training and anomaly detection metrics, the concise suste, can be constructed to its simple equation as shown in \eqref{eq:mse}. The symbol $y$, $\hat{y}$, and $N$ represent the target value, predicted value, and the number of the sample respectively.

\begin{equation}
    L_{MSE}(y, \hat{y}) = \frac{1}{N}\sum_{n=1}^{N}(y - \hat{y})^{2}
    \label{eq:mse}
\end{equation}

However, the other architectures are constructed with a more complexed neural network structure or loss function for training \cite{mse_nguyen2019conad, mse_gong2019memae, mse_schlegl2019fanogan, mse_park2020hpgan}. For example, a generative adversarial network can replace a simple auto-encoder alternatively. However, an additional adversarial loss function is needed to utilizing adversarial learning and it occurs increasing computational cost in the above context.

For sophistic learning purposes in previous studies, varied loss functions and regularization techniques are proposed to support the garget loss convergence better such as winner-takes-all loss, sparsity regularization, and \emph{izi} encoder training. In order to utilize such a technique as above, hard efforts are required for numerous theory, appropriately matching to thier neural networks, establishment and verification processes. In addition, the computational cost increases in the same context as the case of the neural network structure advancement.

\subsection{Logarithmic methods}
\label{subsec:dl_anomaly_detection}
In prior researches, Vincent Arsigny et al. show the effectiveness of the log scaling method \cite{log_arsigny2005log, log_arsigny2006log}. The logarithmic method facilitates joint estimation and smoothing largely with almost the very close cost of the original Euclidean distance on their research.

In another research, Tianjia He et al. have trained a neural network with three losses MSE, mean squared logarithmic error (MSLE), and mean absolute error respectively to compare the performance in the anomaly detection task \cite{log_he2020log}. The MSLE and MAE that used in thier study are shown in \eqref{eq:msle} and \eqref{eq:mae} sequantially.

\begin{equation}
    \begin{aligned}
        L_{MSLE}(y, \hat{y}) = \frac{1}{N}\sum_{n=1}^{N}\Bigl(\ln(y + 1)-\ln(\hat{y} + 1)\Bigr)^{2}
    \end{aligned}
    \label{eq:msle}
\end{equation}

\begin{equation}
    L_{MAE}(y, \hat{y}) = \frac{1}{N}\sum_{n=1}^{N}|y - \hat{y}|
    \label{eq:mae}
\end{equation}

The above research shows the advantage of MSE and MSLE than MAE by comparing reconstruction loss between normal and anomalies after training. Also, they calrify that MSLE shows the better performance than MSE and MAE cases. However, they does not shows the robustness of loss function within training process.

\section{Proposed approach}
\label{sec:proposed_approach}
In this section, the approach to enhance the loss convergence in neural network training is described. Moreover, the theoretical comparison between baseline and proposed function is presented.

\subsection{Logarithmic mean squared error}
\label{subsec:logarithmic_mse}
In this paper, the loss function is proposed for training the generative neural network for the anomaly detection tasks. The proposed loss function is constructed based on MSE as shown in \eqref{eq:mse}.

The MSLE function, an MSE-based logarithmic loss function, can be used. However, the MSLE function cannot be concluded that the design is advantageous for gradient descent as the loss convergence because the logarithmic scalling is applied to the target and output values, not applied to loss value itself. Thus, the log scaling is simply applied to loss calue directly in this study with the direction of maintaining the structure of the MSE function as shown in \eqref{eq:lmse}. The proposed loss function is defined with a name logarithmic mean squared error (LMSE).

\begin{equation}
    \begin{aligned}
        L_{LMSE}(y, \hat{y}) = \frac{1}{N}\sum_{n=1}^{N}-\log\Bigl(1 - \frac{(y - \hat{y})}{\max(y - \hat{y})} \Bigr)
    \end{aligned}
    \label{eq:lmse}
\end{equation}

When the range of the value $y$ and $\hat{y}$ are regularized between 0 and 1, the range of the $(y-\hat{y})^{2}$ with log-scaling is $-\infty$ to 0. To apply the MSE function with log-scaling for the loss minimizing task, as same as optimization, the range of the loss should be adjusted as 0 to $C$ (positive finate value) form.

For utilizing the logarithmic function, the negative form is needed to flip upside down the representation range of the logarithmic function. After the above, the output range of the function will be changed as $\infty$ to 0. Then, the $1-x$ form is applied sequentially to flip starting and ending values. The MSE value $(y-\hat{y})^{2}$ is applied to the $x$ in the above form.

However, after conducting the above procedure, the ending point of the loss function is directed $\infty$. Thus, the small floating-point value $\epsilon$ is added to the $1-x$ form for easing the ending point to the value $C$. Finally, the range of the LMSE function is 0 to $C$.

\subsection{Optimization}
\label{subsec:optimization}
For the neural network optimization, the gradient descent method is utilized that can be summarized as shown in \eqref{eq:update}. The symbol $\theta$ means parameter of the neural network, and the $L$ means the target loss function.

\begin{equation}
    \theta^{(t+1)} = \theta^{(t)} - \eta\nabla{L^{(t)}}
    \label{eq:update}
\end{equation}

The loss $L^{(t)}$ is calculated at $t$-th update in training iterations. Also, the gradient of the loss, $\nabla{L^{(t)}}$, is derived after the above. In the derivative surface that derived from the loss function, if the area closer to the target point is flatter, and the farther point is steep, the gradient descent method will be effectively utilized.

Before applying neural network training, the derivatives of the two-loss functions, MSE and LMSE are compared. The derivative of MSE and LMSE are shown in \eqref{eq:diff_mse} and \eqref{eq:diff_lmse} respectively. For more detail, it can be dealt with additional derivative including the chain rule by the parameters and input of the neural network. However, this paper only deals with the loss function-level derivatives. Because, the derivative that focused on parameter or input in front of loss function is same between MSE and LMSE on the same neural network.

\begin{equation}
    \begin{aligned}
        &\nabla{L_{MSE}} = \frac{\partial{L_{MSE}}}{\partial{\hat{y}}} = \frac{\partial{}}{\partial{\hat{y}}}\biggl(\frac{1}{N}\sum_{n=1}^{N}(y - \hat{y})^{2}\biggr) \\
        &= \frac{1}{N}\sum_{n=1}^{N}(-2y + 2\hat{y})
    \end{aligned}
    \label{eq:diff_mse}
\end{equation}

The MSE case shows simple derivative as adding form with the two values. However, in the LMSE case, the derivative is composited with additional scaling factor as $1-(y-\hat{y})^2$ form, than MSE case. 

\begin{equation}
    \begin{aligned}
    &\nabla{L_{LMSE}} = \frac{\partial{L_{LMSE}}}{\partial{\hat{y}}} \\
    &= \frac{\partial{}}{\partial{\hat{y}}}\frac{1}{N}\sum_{n=1}^{N}-\ln\Bigl(1 - (y - \hat{y})^{2}\Bigr) \\
    &= \frac{1}{N}\sum_{n=1}^{N}\Biggl(\frac{\partial{}}{\partial{f}}-\ln{f}\Biggr)
    \Biggl(\frac{\partial{}}{\partial{\hat{y}}}\Bigl(1 - (y - \hat{y})^{2}\Bigr)\Biggr) \\
    &= \frac{1}{N}\sum_{n=1}^{N}\frac{-2y + 2\hat{y}}{1 - (y - \hat{y})^{2}}
    \end{aligned}
    \label{eq:diff_lmse}
\end{equation}

\begin{figure}[t]
    \begin{center}
        \includegraphics[width=0.90\linewidth]{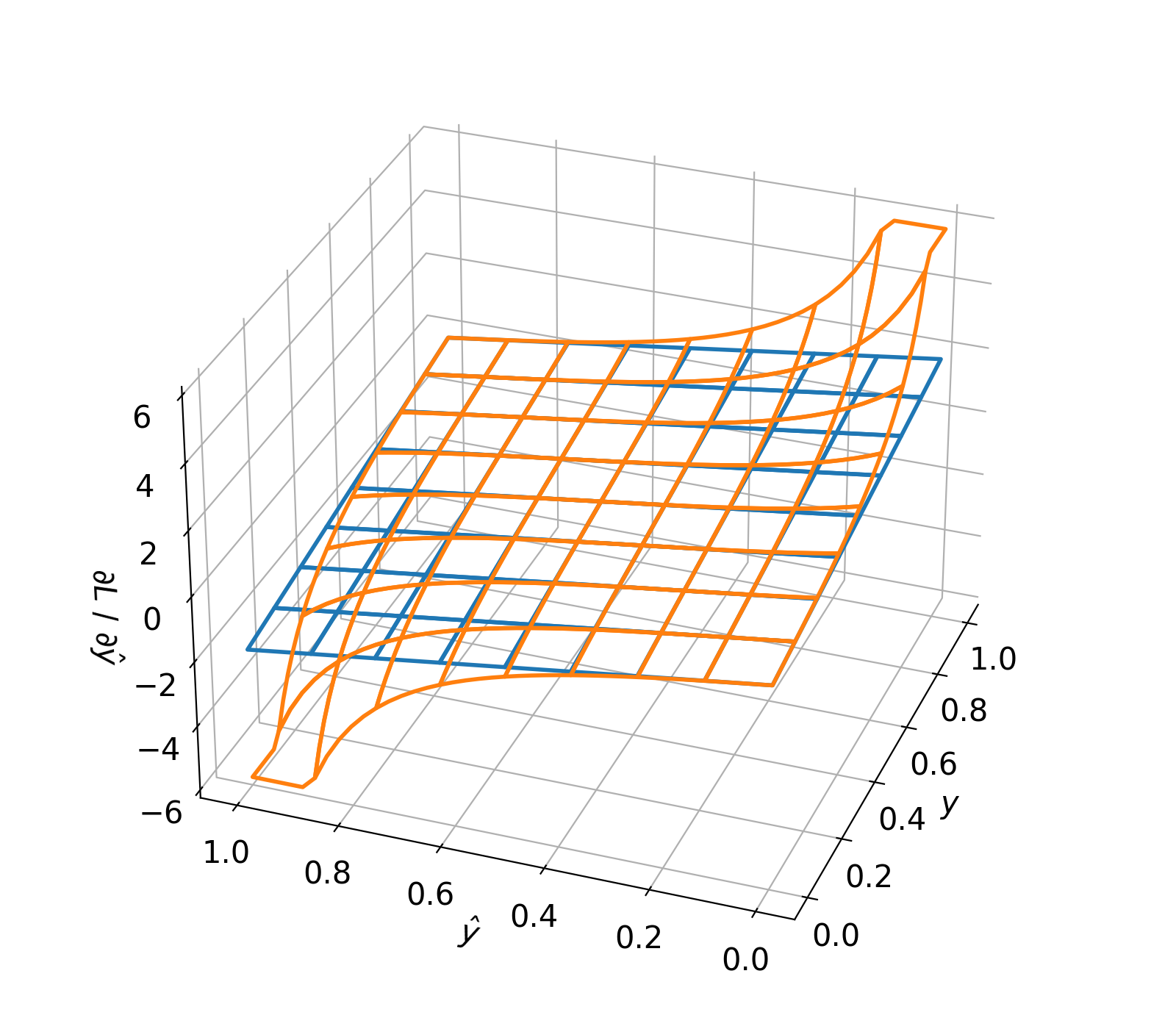}
    \end{center}
    \vspace*{-5mm}
    \caption{The gradient surfaces of each loss surface are shown in Fig.~\ref{fig:surface}. The blue and orange colored surfaces represent the gradient of MSE and LMSE respectively. For the convenience of visualization, when the value of the gradient exceeds a certain level, it is expressed by a clipped value.}
    \label{fig:surface_deriv}
\end{figure}

To compare the above two derivatives, the surfaces are shown in Fig.~\ref{fig:surface} and Fig.~\ref{fig:surface_deriv}. In each figure, blue and orange color represents surface of MSE and LMSE respectively. The LMSE surface, shows a steeper slope when the gap between the target $y$ and the predicted value $\hat{y}$ increases. The above characteristic helps the parameter update stably toward the minimum loss point following the gradient descent rule.

\section{Experiments}
\label{sec:experiments}
\subsection{Environments}
\label{subsec:environments}
In this section, the anomaly detection tasks are conducted by applying the two-loss functions, MSE and LMSE, respectively. Both two-loss functions are applied to the same simple convolutional auto-encoder (CAE). The CAE is constructed static channel expansion as \emph{16, 32, 64} sequentially in encoder module. The decoder module has reversed form with the encoder module.

The two popular datasets, MNIST and Fashion-MNIST, are used for the experiment \cite{data_lecun1998mnist, data_xiao2017fmnist}. The above two dataset has 10 classes respectively. The anomaly detection experiments are designed in which considering one of the 10 classes as normal and the others as abnormal for each dataset. Thus, each dataset will be transformed to the 10 anomaly detection tasks. 

To clarify the superiority of the proposed method by comparison, the one of hyperparameter searching algorithms can be considered \cite{tune_larochelle2007grid, tune_bergstra2012random}. In this paper, the grid search method is adopted to compare the performance by loss function at equivalent hyperparameter. The two hyperparameters, dimensions of the latent vector and the learning rate, are set for tuning in this paper. The kernel size is set with static value \emph{3} for the convolutional filter.

The area under the receiver operating characteristic curve (AUROC) is used as a anomaly detection performance indicator for comparison \cite{auroc_fawcett2006auroc}.

\subsection{Results}
\label{subsec:results}

\begin{figure}[t]
    \begin{center}
        \begin{tabular}{cc}
		    \includegraphics[width=0.48\linewidth]{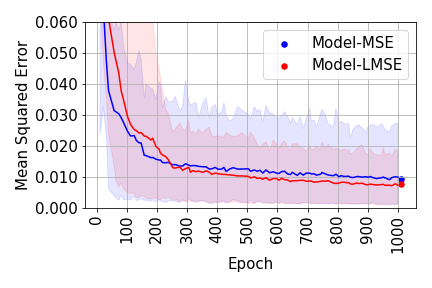}
		    \includegraphics[width=0.48\linewidth]{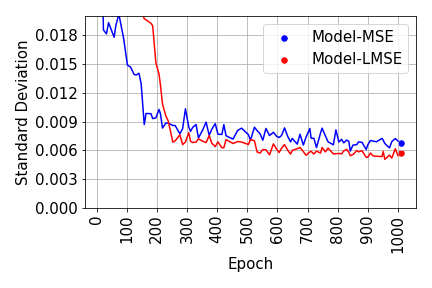}
		\end{tabular}
	\end{center}
	\vspace*{-5mm}
	\caption{Loss convergence when training MNIST dataset. The left graph shows loss range for each epoch with varied hyperparameter. Also, the left graph shows averaged loss with the bold line. In the overall view, the LMSE case performs better loss convergence. Also, the standard deviation of the LMSE case is better than MSE case.}
	\label{fig:loss_mnist}
\end{figure}

\begin{table}[t]
    \centering
    \caption{The anomaly detection performance in the MNIST dataset\\}
    \begin{tabular}{lcc}
        \hline
            \textbf{Normal} & \textbf{MSE} & \textbf{LMSE} \\
        \hline
            0 & 0.981 $\pm$ 0.007 & \textbf{0.983} $\pm$ \textbf{0.006} \\ 
            1 & 0.983 $\pm$ 0.029 & \textbf{0.989} $\pm$ \textbf{0.011} \\ 
            2 & 0.837 $\pm$ 0.068 & \textbf{0.852} $\pm$ \textbf{0.051} \\ 
            3 & 0.881 $\pm$ 0.028 & \textbf{0.888} $\pm$ \textbf{0.022} \\ 
            4 & 0.882 $\pm$ 0.038 & \textbf{0.886} $\pm$ \textbf{0.034} \\ 
            5 & 0.823 $\pm$ 0.128 & \textbf{0.836} $\pm$ \textbf{0.116} \\ 
            6 & \textbf{0.938} $\pm$ 0.043 & 0.937 $\pm$ \textbf{0.036} \\ 
            7 & 0.922 $\pm$ 0.032 & \textbf{0.927} $\pm$ \textbf{0.028} \\ 
            8 & 0.817 $\pm$ 0.037 & \textbf{0.834} $\pm$ \textbf{0.033} \\ 
            9 & 0.903 $\pm$ 0.050 & \textbf{0.909} $\pm$ \textbf{0.046} \\ 
        \hline
            Total & 0.887 $\pm$ 0.074 & \textbf{0.893} $\pm$ \textbf{0.067} \\ 
        \hline
    \end{tabular}
    \label{table:performance_mnist}
\end{table}

Before comparing the anomaly detection performance, The loss convergence graphs in training procedure are summarized in Fig.~\ref{fig:loss_mnist}. In comparing the loss convergence, the LMSE case shows a more stable form at least in the MNIST dataset. In the LMSE case, both the averaged loss and the standard deviation of the loss are lower, as same as better, than MSE case.

In the view of the AUROC, the average performance of the log-scaled MSE case is mesuared 0.006 higher than the MSE case for 10 of anomaly detection tasks in MNIST dataset. The performance detail for each task is shown in Table~\ref{table:performance_mnist}. Referring to Table~\ref{table:performance_mnist}. the performance gain is measured up to 0.017 by adopting LMSE.

\begin{figure}[t]
    \begin{center}
        \begin{tabular}{cc}
		    \includegraphics[width=0.48\linewidth]{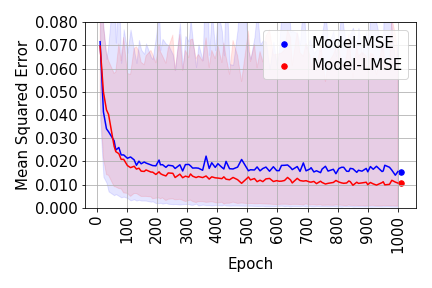}
		    \includegraphics[width=0.48\linewidth]{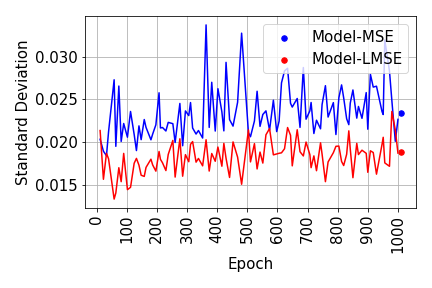}
		\end{tabular}
	\end{center}
	\vspace*{-5mm}
	\caption{Loss convergence when training Fashion-MNIST dataset. The pattern of the convergence is differ from Fig.~\ref{fig:loss_mnist}, but the whole context is almost same.}
	\label{fig:loss_fmnist}
\end{figure}

The training procedure with the Fashion-MNIST datset is shown in Fig.~\ref{fig:loss_fmnist}. The pattern of loss convergence is slightly different from the case of MNIST. However, the context in which loss convergence is better in the LMSE case than MSE is the same as before.

\begin{table}[t]
    \centering
    \caption{The anomaly detection performance in the Fashion-MNIST dataset\\}
    \begin{tabular}{lcc}
        \hline
            \textbf{Normal} & \textbf{MSE} & \textbf{LMSE} \\
        \hline
            T-shirt/top & \textbf{0.883} $\pm$ \textbf{0.011} & 0.882 $\pm$ 0.012 \\ 
            Trouser & \textbf{0.987} $\pm$ \textbf{0.005} & 0.986 $\pm$ 0.006 \\ 
            Pullover & \textbf{0.867} $\pm$ \textbf{0.015} & 0.866 $\pm$ \textbf{0.015} \\ 
            Dress & \textbf{0.910} $\pm$ \textbf{0.008} & 0.908 $\pm$ 0.010 \\ 
            Coat & \textbf{0.903} $\pm$ \textbf{0.009} & 0.901 $\pm$ 0.012 \\ 
            Sandal & \textbf{0.874} $\pm$ \textbf{0.020} & 0.867 $\pm$ 0.023 \\ 
            Shirt & 0.794 $\pm$ \textbf{0.013} & \textbf{0.797} $\pm$ 0.014 \\ 
            Sneaker & \textbf{0.979} $\pm$ \textbf{0.004} & \textbf{0.979} $\pm$ \textbf{0.004} \\ 
            Bag & \textbf{0.862} $\pm$ 0.034 & 0.857 $\pm$ \textbf{0.017} \\ 
            Ankle boot & 0.973 $\pm$ \textbf{0.008} & \textbf{0.974} $\pm$ 0.009 \\ 
        \hline
            Total & \textbf{0.903} $\pm$ \textbf{0.060} & 0.901 $\pm$ \textbf{0.060} \\ 
        \hline
    \end{tabular}
    \label{table:performance_fmnist}
\end{table}

In the quantitative comparison with Fashion-MNIST dataset, the MSE case is 0.002 higher than LMSE case that differ from the MNIST dataset. In Fashion-MNIST, even when the performance of the MSE case is better but the difference is insignificant and it is negligible, referring to the fact in MNIST experiments that the LMSE performance is higher than the MSE case and the difference was relatively large. If the performance is in similar level, using LMSE is better for stable loss convergence.

\section{Conclusion}
\label{sec:conclusion}
As the learning progressed, the degree of loss convergence of LMSE case is better than MSE case. Also, the standard deviation according to the varied hyperparameter in training iterations is lower in LMSE case which means more stable.

In terms of anomaly detection performance, the LMSE case on the MNIST dataset shows 0.006 higher AUROC, and 0.007 lower standard deviation. In the Fashion-MNIST dataset, the MSE case showed a 0.002 higher performance. However, the MSE and LMSE show the equivalent level in terms of performance stability.

In overall view, the LMSE function is recommended based on robust loss convergence. Also, the above robustness will utilize high probability for achieving better performance in anomaly detection task.



\pagebreak
\begin{appendices}
\section{Detail of LMSE}

Another important factor along with differentiability that shown in \eqref{eq:diff_lmse} is preserving the desirable properties of the MSE as a loss function.
In the problem of linear optimization, there are essential requirements of the loss function, such as the convexity and the existence of a closed-form solution.
Due to the nonlinear optimization like neural network training, the loss surface is in the form of multiple local minima but this study supposes them as an absolute convex to prove that the LMSE does not impair the inherent characteristics of the MSE.

First, the proof for the convexity of LMSE will be shown.
To show the convexity of LMSE, the $x$ and $y$ are defined as a single input and target data respectively.
For batch processing, $x$ and $y$ are represented as $X$ and $Y$ by size $N$ (number of samples).
The symbol $W$ and $f$ represent the parameter (weight) and the number of features.
The dimension of each symbol is following.

\begin{itemize}
    \item $y \in \mathbb{R}^{hwc},\quad{} x \in \mathbb{R}^{f},\quad{} W \in \mathbb{R}^{f \times hwc}$
    \item $Y \in \mathbb{R}^{N \times hwc},\quad{} X \in \mathbb{R}^{N \times f}$
\end{itemize}

The MSE and LMSE is represented with matrix multiplication form as \eqref{eq:l2_convex} and \eqref{eq:ll2_convex}.

\begin{equation}
    \begin{aligned}
        L_{MSE}(W) &= \sum^{N}_{n} (y^{(n)} - W^{T}{x^{(n)}})^{2} \\
        &= (Y-XW)^{T}(Y-XW) \\
        &= Y^{T}Y - 2Y^{T}XW + W^{T}X^{T}XW
    \end{aligned}
    \label{eq:l2_convex}
\end{equation}

\begin{equation}
    \begin{aligned}
        L_{LMSE}(W) &= \sum^{N}_{n}-\log\Bigl(1 - (y^{(n)}-\hat{y}^{(n)})^{2} \Bigr)\\
        &= -log(I-(Y-XW)^{T}(Y-XW))
    \end{aligned}
    \label{eq:ll2_convex}
\end{equation} 

In the above equations, three terms $Y^{T}Y$, $-2Y^{T}XW$, and $W^{T}X^{T}XW$ are corresponding to the constant, straight line, and curve respectively.
Convexity means that when the two points on the curve are selected randomly, all values between them have a lower or equal value.
Thus, only the curve term is needed to deal with, and if it satisfies the positive semi-definite (PSD), the loss function can be defined as a convex.
The inner product $X^{T}X$ in the curve term is called a gram matrix that satisfies the PSD so transformation by $W$ still satisfied PSD.

To prove the LMSE function is still convex as MSE, the LMSE is decomposed to the inside function $1-L_{MSE}$ and outside function $-log$ as shown in \eqref{eq:flip_concave} and \eqref{eq:logs_convex}.

\begin{equation}
    L_{LMSE_{inside}}(W) = I - L_{MSE}(W)
    \label{eq:flip_concave}
\end{equation} 

\begin{equation}
    L_{LMSE_{outside}}(W) = -log(L_{LMSE_{inside}}(W))
    \label{eq:logs_convex}
\end{equation} 

Each method only conducts changing the sign, bias, or scale.
In the case of \eqref{eq:flip_concave}, curve term $W^{T}X^{T}XW$ is changed to a concave with satisfying negative semi-definite (NSD) since subtraction from the identity matrix.
The above concave form is changed back to the convex by applying \eqref{eq:logs_convex}.

For a more detailed interpretation, the \eqref{eq:logs_convex} can be expressed as a Taylor expansion as shown in \eqref{eq:log_taylor}.
The numerator $(-1)^{n} (x-1)^{n}$ reverses the characteristics of input matrix (PSD or NSD) that includes only values between 0 and 1.
Thus, concave form due to \eqref{eq:flip_concave} step will become convex in all $n$-th terms of Taylor expansion.
In conclusion, it is clearly demonstrated that the convexity of the MSE is preserved even when the LMSE is used.

\begin{equation}
    \begin{aligned}
        -log(x) &= -\sum_{n}^{\infty} \frac{(-1)^{n-1} (x-1)^{n}}{n} \\
        &= \sum_{n}^{\infty} \frac{(-1)^{n} (x-1)^{n}}{n}
        \label{eq:log_taylor}
    \end{aligned}
\end{equation} 

Another desirable property of the loss function like MSE is whether it provides a closed-form solution.
When the loss function has a closed-form solution, the optimization process guarantees to find the global minima.
Before confirming existent of the closed-form solution, differentiated of LMSE function can be derived by three partial functions $a=(Y-XW)^{T}(Y-XW)$, $b=I-a$, and $c=-log(b)$ as \eqref{eq:fliplog_diff}.

\begin{equation}
    \begin{aligned}
        \frac{\delta}{\delta{W}}L_{LMSE}(W) &= \frac{\partial{L_{LMSE}}}{\partial{c}} \frac{\partial{c}}{\partial{b}} \frac{\partial{b}}{\partial{a}} \frac{\partial{a}}{\partial{W}} \\
        &= \frac{-2Y^{T}X + X^{T}XW}{I-(Y-XW)^{T}(Y-XW)}
        \label{eq:fliplog_diff}
    \end{aligned}
\end{equation}

Then, to show the preservation of closed-form solution, the equation is derived by set $\frac{\delta}{\delta{W}}L_{LMSE}(W^{*}) = 0$ as shown in \eqref{eq:fliplog_closed} refer to \eqref{eq:fliplog_diff}.

\begin{equation}
    \begin{aligned}
        \frac{\delta}{\delta{W}}L_{LMSE}(W^{*}) &= 0 \\
        0 &= \frac{-2Y^{T}X + X^{T}XW^{*}}{I-(Y-XW^{*})^{T}(Y-XW^{*})} \\
        0 &= -2Y^{T}X + X^{T}XW^{*} \\
        W^{*} &= (X^{T}X)^{-1}2Y^{T}X
        \label{eq:fliplog_closed}
    \end{aligned}
\end{equation} 

In \eqref{eq:fliplog_closed}, the closed-form solution $W^{*}$ is derived. 
Moreover, the solution $W^{*}$ is the same as the MSE by inheritance.
Through the proof in this section, it becomes clear that even when the gradient changes as a logarithmic scaled function, LMSE, the inherent convexity and optimal solution of the MSE are preserved.

\section{Generalization of LMSE}

The proposed loss transformation approach can be presented in a generalized form that is not limited to MSE.
The generalized expression is shown as \eqref{eq:distance_fliplog}, named Flipping and Log-scaling (FL) loss, and will be discussed in a later study. 
Refer to that, the \textit{flipping} and \textit{log-scaling} steps are represented with partial functions $1-L$ and $-log$.

\begin{equation}
    \begin{aligned}
        L_{FL}(y, \hat{y}) = -\log\Bigl(1 - L(y, \hat{y})\Bigr)
    \end{aligned}
    \label{eq:distance_fliplog}
\end{equation}

To ensure that the input to the log function is necessarily positive, a scaling trick can be used to normalize the input before \eqref{eq:distance_fliplog} to between 0 and 1, as shown in \eqref{eq:scale_trick}.

\begin{equation}
    \begin{aligned}
        L^{\prime}(y, \hat{y}) = \frac{L(y, \hat{y})}{\max{(L(y, \hat{y}))}}(1-\epsilon)
    \end{aligned}
    \label{eq:scale_trick}
\end{equation}

The method adopts a simple approach that only divides all pixel-wise distance by the maximum distance of $L(y, \hat{y})$.
Also, the scaling trick includes a coefficient multiplying operation to the divided value with $(1-\epsilon)$.
By applying the described trick before feeding input into \eqref{eq:distance_fliplog}, we can safely apply FL loss for any loss function while preventing such an infinite explosion of the loss.

\end{appendices}

\end{document}